\documentclass[10pt]{article}

\usepackage[affil-it]{authblk}

\usepackage{cite}
\usepackage{amsmath,amssymb,amsfonts}
\usepackage{algorithmic}
\usepackage{graphicx}
\usepackage{textcomp}
\usepackage{subcaption}
\usepackage{caption}
\usepackage{algorithm}

\providecommand{\keywords}[1]{\textbf{Keywords:} #1}

\begin{document}

\title{A Parts Based Registration Loss for Detecting Knee Joint Areas}
\author{Juha Tiirola%
\thanks{Email: \texttt{juha.a.tiirola@gmail.com}}}
\affil{Research Unit of Mathematical Sciences, University of Oulu, Finland}
\date{\today}

\maketitle

\begin{abstract}
In this paper, a parts based loss is considered for finetune registering knee joint areas. Here the parts are defined as abstract feature vectors with location and they are automatically selected from a reference image. For a test image the detected parts are encouraged to have a similar spatial configuration than the corresponding parts in the reference image.
\end{abstract}


\section{Introduction}
Neural networks based general object detection methods need lot of supervision in the form of bounding box annotations. Some well-known methods in this direction are Fast R-CNN \cite{Girshick}, Faster R-CNN \cite{RenHe} and Yolo \cite{redmon_farhardi}. In contrast to general object detection, in the knee joint area detection problem from bilateral PA fixed flexion X-ray images the amount of variation between samples is small since the pose is shared between images and due to human knee anatomy, there are no big deformations between samples. 

In \cite{tiirola}, the detection of knee joint areas was considered based on template matching. There the registration loss was based on the use of the  normalized cross-correlation. In contrast to \cite{tiirola}, in this paper the detection is based on points matching. More precisely, the points here are abstract feature vectors. By parts we refer to abstract feature vectors equipped with location and which are extracted from a reference image. The term 'part' was used for instance in \cite{Kortylewski0WZY20} where a dictionary of parts was learned by clustering feature map vectors. 
 In \cite{PAYER2019207}, landmark localization was considered using convolutional neural networks where spatial configuration was integrated into heatmap regression. In \cite{Tompson}, human pose was estimated using a convolutional network where the architecture could exploit structural domain constraints such as geometric relationships between body joint locations. We use a similar idea to deal with the parts. Since the camera pose between different bilateral fixed flexion knee X-ray images is shared, the parts should be such that there is a common pattern between the locations of the parts between different images. More precisely, given a test image, the extracted patch from the test image should be such that each part in the extracted patch is near the corresponding part in the reference image and in addition the content of each part in the extracted patch looks like the corresponding content in the reference image. The term 'looks like that' was used in \cite{chen} where explainable image classification was considered based on selecting prototypical parts and by making the classification based on how much each prototype is present in the input image.
   
\section{New method}
\begin{figure}
\centering
\begin{subfigure}[b]{0.2\textwidth}
\centering
\includegraphics[width=\textwidth]{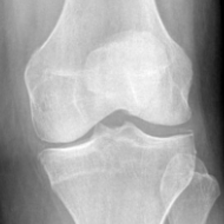}
\caption{$a$}
\end{subfigure}
\hfill
\begin{subfigure}[b]{0.75\textwidth}
\centering
\includegraphics[width=\textwidth]{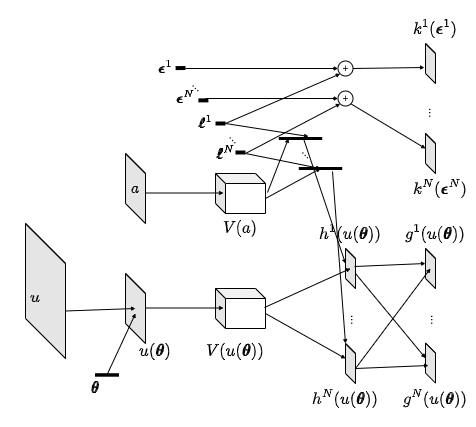}
\caption{Architecture}
\end{subfigure}
\caption{Here $a$ is the manually selected reference image, $N$ is the number of the parts, $u(\pmb{\theta})$ is the extracted patch from the input image $u$ corresponding to the parameter $\pmb{\theta}$. Heatmap $k^i(\pmb{\epsilon}^i)$ is a Gaussian blob centered at $\pmb{\epsilon}^i+\pmb{\ell}^i$. $g^i(u(\pmb{\theta}))$ is obtained by jointly denoising part detection heatmaps $h^1(u(\pmb{\theta})),\dots,h^N(u(\pmb{\theta})).$ In the first phase, $\pmb{\ell}^i, i=1,\dotsc,N$, are learnt using the whole dataset. In the second phase, given a test image $u$, image specific parameters $\pmb{\theta}\in\mathbb{R}^4$ and $\pmb{\epsilon}^i\in\mathbb{R}^2,$ $i=1,\dotsc,N$, are optimized by minimizing energy~\eqref{eq:u_test}.}
\label{fig:architecture}
\end{figure}
The reference image $a$ and the architecture of the new method are shown in Figure~\ref{fig:architecture}. 
We first select a reference image $a$ manually. Given a test image $u$, we try to extract a patch from $u$ which resembles $a$ as much as possible. We formulate this similarity using parts. We use the VGG16 network \cite{Simonyan} pretrained on the ImageNet as the backbone network. We freeze  the weights of the backbone. We denote by $V\in\mathbb{R}^{28\times28\times512}$ the feature tensor from the 22th layer of the VGG16  network followed by normalization. Thus, $||V_{r,s,:}||_2=1$ for all $r,s$.

We first determine the parts from the reference image utilizing the whole dataset. In the $[-1,1]^2$ image coordinate domain, we assume $\pmb{\ell}^i\in[-1,1]^2$, $i=1,\dotsc,N$, denote the centers of the parts in image $a$. We use $N=9$ and initialize the $\pmb{\ell}^i$ as points from a regular grid $(-0.5,-0.5)+r(0.5,0)+s(0,0.5)$ where $r,s\in\{0,1,2\}$.  The locations $\pmb{\ell}^i$  are optimized over a dataset. For each $i$ we interpolate $V(a)$ at $\pmb{\ell}^i$. The resulting vectors are normalized and we get  $V(a)_{\pmb{\ell}^i}\in\mathbb{R}^{512}$ which encodes the content of the part $i$. Thus, $||V(a)_{\pmb{\ell}^i}||_2=1.$ For an image $u$ from the dataset, we define $h^i(u)\in[0,1]^{28\times28}$ by 
\begin{equation}
h^i(u)_{r,s}:=\langle V(u)_{r,s},V(a)_{\pmb{\ell}^i}\rangle_{L^2} \label{eq:heat}
\end{equation}
where $\langle\cdot,\cdot\rangle_{L^2}$ denotes the $L^2$ inner product between two vectors. Thus, $h^i(u)$ is the heatmap which tells how much in $u$ there is part $i$ in each location. The locations are optimized over the whole dataset by solving 
\[
\min_{\{\pmb{\ell}^i\}_{i=1}^N}\sum_u\sum_{i=1}^N ||h^i(u)||_1+\lambda(1-\max_{r,s}h^i(u)_{r,s})
\]
where we use $\lambda=0.1$. Ideally, the locations should be such that in an image $u$ the part $i$ is detected in a single location. In the above minimization the first term encourages sparsity in $h^i(u)$ and the second loss encourages that the maximum value in $h^i(u)$ is near $1$. After the optimization, we freeze the locations $\{\pmb{\ell}^i\}_{i=1}^N.$ 

The second step is to extract a patch from each test image $u$ such that the extracted patch resembles $a$ visually as much as possible. This optimal extraction is formulated as an optimization problem. There are $4+2N$ variables in the minimization problem. The first four variables are transformation parameters $\pmb{\theta}\in\mathbb{R}^4$ for the spatial transformer \cite{NIPS2015_33ceb07b} which determine the extracted patch from the input image. The extracted patch is denoted by $u(\pmb{\theta})$. The rest of the optimization variables are disturbance vectors. Since there are slight deformations between knee images, ideally the location of the part $i$ in a well registered test image is the location of the part in the reference image plus some small disturbance vector. This disturbance vector is denoted by $\pmb{\epsilon}^i$ for part $i$.

The heatmaps $h^i(u(\pmb{\theta}))$ obtained using \eqref{eq:heat} denote the detected parts $i$ in the patch $u(\pmb{\theta})$. These heatmaps $h^i$ are noisy and ambiguous. We use a  method similar to \cite{PAYER2019207} and \cite{Tompson} to denoise the parts heatmaps $h^j$ jointly by utilizing the spatial configuration between the locations of the parts. If image $v$ resembles the reference image $a$ and if there is a peak in $h^i(v)$ near $\pmb{\ell}^i$, then if we move from the peak along the vector $\pmb{\ell}^j-\pmb{\ell}^i$, then near the resulting position there should be a peak in $h^j(v)$. The denoised version of $h^j$ is denoted by $g^j$ and is obtained by $g^j(v)=M_j(v)\odot h^j(v)$ where the multiplier image is      
\[
M_j(v):=\frac{1}{N-1}\sum_{r=1,r\neq j}^N\max_{\overline{k}}\left( G(\overline{k})\, \tau_{\pmb{\ell}^j-\pmb{\ell}^r+\overline{k}}(h^r(v))\right).
\]
Above $\tau$ denotes the translation operator and $G$ is the Gaussian kernel with the standard deviation $0.08$ in $[-1,1]^2$ domain. In the multiplier, $(M_j(v))_{a,b}$ is large when for as many $r$ as possible, $r\neq j$, $h^r(v)$ is large near $(a,b)+\pmb{\ell}^j-\pmb{\ell}^r.$  

We also model the locations of the parts directly. We assume that the location of part $i$ in a well extracted patch from $u$ is $\pmb{\ell}^i+\pmb{\epsilon}^i$ where $||\pmb{\epsilon}^i||$ is small. We place a Gaussian blob with the standard deviation $0.08$ at the location $\pmb{\ell}^i+\pmb{\epsilon}^i$. The heatmap obtained in this way is denoted by $k^i(\pmb{\epsilon}^i)$. 

The parameters $\pmb{\epsilon}^i$, $\pmb{\theta}$ are found jointly by  solving  
\begin{align}
&\min_{\{\pmb{\epsilon}^i\}_{i=1}^N,\pmb{\theta}} -\sum_{n=1}^N\langle 1+k^n(\pmb{\epsilon}^n),g^n(u(\pmb{\theta}))\odot g^n(u(\pmb{\theta}))\rangle_{L^2}+\lambda\sum_{n=1}^N ||\pmb{\epsilon}^n||_2^2 \nonumber  \\ 
&=\min_{\{\pmb{\epsilon}^i\}_{i=1}^N,\pmb{\theta}}-\sum_{n=1}^N\sum_{r,s}(1+k^n(\pmb{\epsilon}^n)_{r,s}) (g^n(u(\pmb{\theta}))_{r,s})^2+\lambda\sum_{n=1}^N ||\pmb{\epsilon}^n||_2^2. \nonumber \\
&=:\min_{\{\pmb{\epsilon}^i\}_{i=1}^N,\pmb{\theta}} E(u,\pmb{\epsilon}^1,\dotsc,\pmb{\epsilon}^N,\pmb{\theta}). \label{eq:u_test}
\end{align}
The loss to be minimized incorporates the requirement that in a well extracted patch the location of the part $i$ should be close to $\pmb{\ell}^i$ and in the heatmap $h^i$ and $g^i$ there should be a peak near $\pmb{\ell}^i$.

\section{Parametrization}
We transform minimization problem \eqref{eq:u_test} to an unconstrained minimization problem. See also \cite{tiirola} for the interpretation of the parameters. We denote by $(\pmb{w},\pmb{v})\in\mathbb{R}^{2N+4}$ the unconstrained parameter vector. We express $\pmb{\theta}$, $\pmb{\epsilon}^1,\dotsc,\pmb{\epsilon}^N$ as functions of $(\pmb{w},\pmb{v})$. 
For $\{\pmb{\epsilon}^i\}_{i=1}^N$ we use $(\pmb{\epsilon}^1,\dotsc,\pmb{\epsilon}^i,\dotsc,\pmb{\epsilon}^N)=\frac{1}{4}\tanh(\pmb{w})$ and we initialize $\pmb{w}=\pmb{0}\in\mathbb{R}^{2N}$. For the scale we use 
\begin{equation}
\pmb{\theta}_1(\pmb{v}_1)=\left(s_1-\frac{1}{4}\right)+\frac{1}{2}\left(\sigma(\pmb{v}_1)\right)=\left(s_1-\frac{1}{4}\right)+\frac{1}{2}\left(\frac{1}{2}\left(1+\tanh(\pmb{v}_1)\right)\right) \label{eq:scale}
\end{equation}
where $s_1$ is fixed. For the translations we use
\[
\pmb{\theta}_2(\pmb{v})=(1-\pmb{\theta}_1(\pmb{v}_1))(-1+2\sigma(\pmb{v}_2)) \text{ and } \pmb{\theta}_3(\pmb{v})=(1-\pmb{\theta}_1(\pmb{v}_1))(-1+2\sigma(\pmb{v}_3))
\]
and for the rotation we use
\[
\pmb{\theta}_4(\pmb{v}_4)=\frac{1}{10}\tanh(\pmb{v}_4).
\]
We initialize $\pmb{v}=\pmb{0}\in\mathbb{R}^4$.

The transformation matrix corresponding to $\pmb{v}$ is given by
\begin{equation}\label{eq:tr_matrix_A}
A(\pmb{\theta}(\pmb{v})):=\begin{bmatrix}
\pmb{\theta}_1 \cos(\pmb{\theta}_4) & -\pmb{\theta}_1 \sin(\pmb{\theta}_4) & \pmb{\theta}_2 \\
\pmb{\theta}_1 \sin(\pmb{\theta}_4) & \pmb{\theta}_1 \cos(\pmb{\theta}_4) & \pmb{\theta}_3
\end{bmatrix}.
\end{equation}
We transform problem \eqref{eq:u_test} to an unconstrained minimization problem 
\begin{equation}
\min_{\{\pmb{w}_i\}_{i=1}^N,\pmb{v}} E(u,\pmb{\epsilon}^1(\pmb{w}_1),\dotsc,\pmb{\epsilon}^N(\pmb{w}_N),\pmb{\theta}(\pmb{v})). \label{eq:unconst}
\end{equation}
\section{Experimental results}
We test the minimization of \eqref{eq:unconst} for several scales $s_1$ in \eqref{eq:scale}. We do the minimization for all $s_1\in\{0.65,0.70,0.75,\dotsc,1.15,1.20\}$ and we take the patch corresponding to the smallest overall loss value. We use $\lambda=10^{-6}$ in equation~\eqref{eq:u_test}.

We only consider finetune registration. We use as inputs $u$ randomly slightly enlargened outputs of the neural method \cite{tiirola} since our goal is to test if there is a visual correspondence between the reference image $a$ and the minimizer of \eqref{eq:unconst}.

\begin{figure*}
\center
\begin{tabular}{c|c|c}
\begin{subfigure}{0.13\textwidth}
\includegraphics[width=0.95\textwidth]{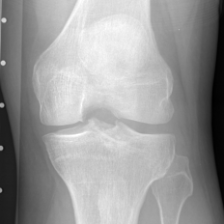}
\end{subfigure}
\begin{subfigure}{0.13\textwidth}
\includegraphics[width=0.95\textwidth]{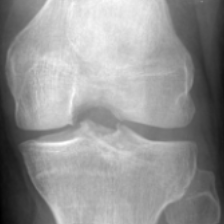}
\end{subfigure}
&
\begin{subfigure}{0.13\textwidth}
\includegraphics[width=0.95\textwidth]{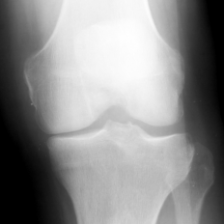}
\end{subfigure}
\begin{subfigure}{0.13\textwidth}
\includegraphics[width=0.95\textwidth]{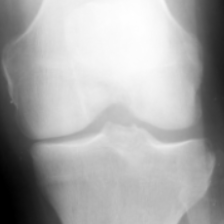}
\end{subfigure}
&
\begin{subfigure}{0.13\textwidth}
\includegraphics[width=0.95\textwidth]{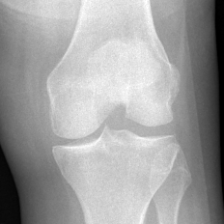}
\end{subfigure}
\begin{subfigure}{0.13\textwidth}
\includegraphics[width=0.95\textwidth]{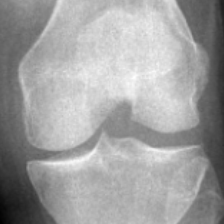}
\end{subfigure}
\\
\begin{subfigure}{0.13\textwidth}
\includegraphics[width=0.95\textwidth]{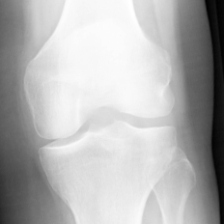}
\end{subfigure}
\begin{subfigure}{0.13\textwidth}
\includegraphics[width=0.95\textwidth]{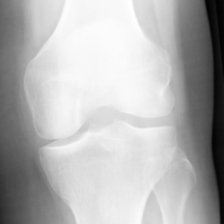}
\end{subfigure}
&
\begin{subfigure}{0.13\textwidth}
\includegraphics[width=0.95\textwidth]{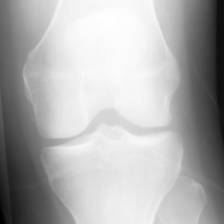}
\end{subfigure}
\begin{subfigure}{0.13\textwidth}
\includegraphics[width=0.95\textwidth]{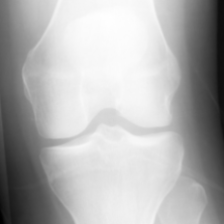}
\end{subfigure}
&
\begin{subfigure}{0.13\textwidth}
\includegraphics[width=0.95\textwidth]{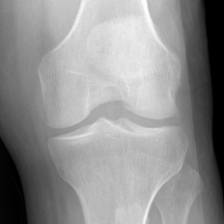}
\end{subfigure}
\begin{subfigure}{0.13\textwidth}
\includegraphics[width=0.95\textwidth]{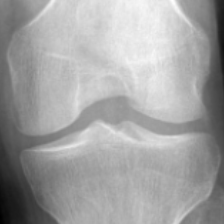}
\end{subfigure}
\\
\begin{subfigure}{0.13\textwidth}
\includegraphics[width=0.95\textwidth]{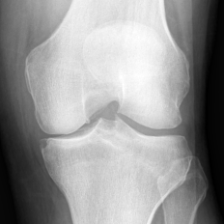}
\end{subfigure}
\begin{subfigure}{0.13\textwidth}
\includegraphics[width=0.95\textwidth]{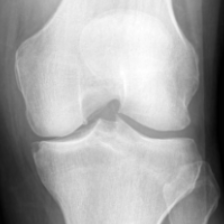}
\end{subfigure}
&
\begin{subfigure}{0.13\textwidth}
\includegraphics[width=0.95\textwidth]{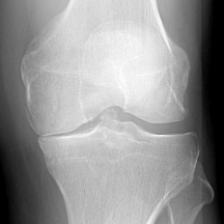}
\end{subfigure}
\begin{subfigure}{0.13\textwidth}
\includegraphics[width=0.95\textwidth]{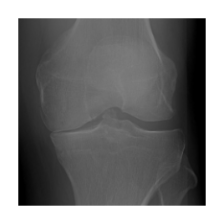}
\end{subfigure}
&
\begin{subfigure}{0.13\textwidth}
\includegraphics[width=0.95\textwidth]{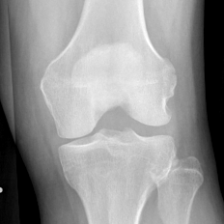}
\end{subfigure}
\begin{subfigure}{0.13\textwidth}
\includegraphics[width=0.95\textwidth]{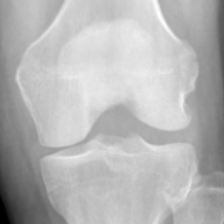}
\end{subfigure}
\\
\begin{subfigure}{0.13\textwidth}
\includegraphics[width=0.95\textwidth]{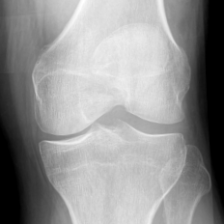}
\end{subfigure}
\begin{subfigure}{0.13\textwidth}
\includegraphics[width=0.95\textwidth]{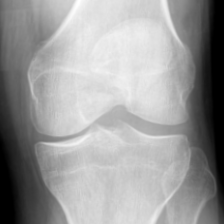}
\end{subfigure}
&
\begin{subfigure}{0.13\textwidth}
\includegraphics[width=0.95\textwidth]{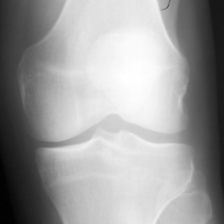}
\end{subfigure}
\begin{subfigure}{0.13\textwidth}
\includegraphics[width=0.95\textwidth]{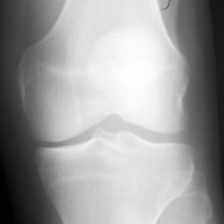}
\end{subfigure}
&
\begin{subfigure}{0.13\textwidth}
\includegraphics[width=0.95\textwidth]{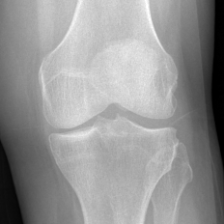}
\end{subfigure}
\begin{subfigure}{0.13\textwidth}
\includegraphics[width=0.95\textwidth]{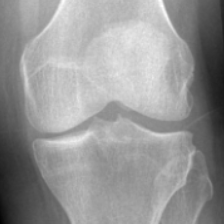}
\end{subfigure}
\\
\begin{subfigure}{0.13\textwidth}
\includegraphics[width=0.95\textwidth]{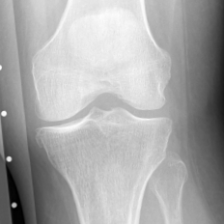}
\end{subfigure}
\begin{subfigure}{0.13\textwidth}
\includegraphics[width=0.95\textwidth]{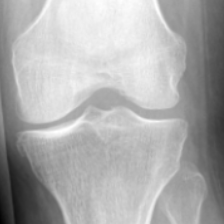}
\end{subfigure}
&
\begin{subfigure}{0.13\textwidth}
\includegraphics[width=0.95\textwidth]{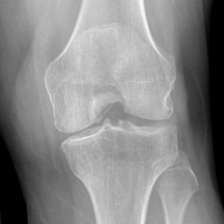}
\end{subfigure}
\begin{subfigure}{0.13\textwidth}
\includegraphics[width=0.95\textwidth]{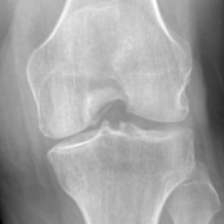}
\end{subfigure}
&
\begin{subfigure}{0.13\textwidth}
\includegraphics[width=0.95\textwidth]{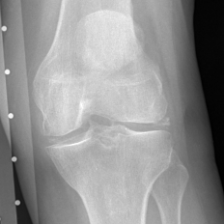}
\end{subfigure}
\begin{subfigure}{0.13\textwidth}
\includegraphics[width=0.95\textwidth]{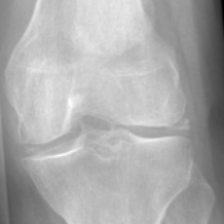}
\end{subfigure}
\\
\begin{subfigure}{0.13\textwidth}
\includegraphics[width=0.95\textwidth]{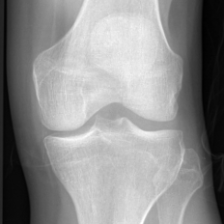}
\end{subfigure}
\begin{subfigure}{0.13\textwidth}
\includegraphics[width=0.95\textwidth]{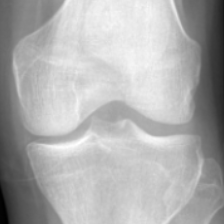}
\end{subfigure}
&
\begin{subfigure}{0.13\textwidth}
\includegraphics[width=0.95\textwidth]{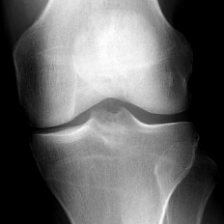}
\end{subfigure}
\begin{subfigure}{0.13\textwidth}
\includegraphics[width=0.95\textwidth]{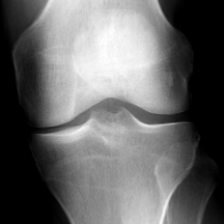}
\end{subfigure}
&
\begin{subfigure}{0.13\textwidth}
\includegraphics[width=0.95\textwidth]{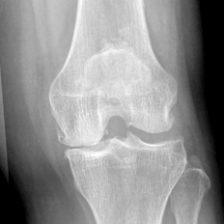}
\end{subfigure}
\begin{subfigure}{0.13\textwidth}
\includegraphics[width=0.95\textwidth]{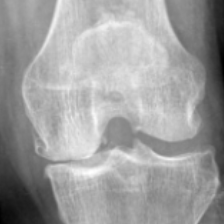}
\end{subfigure}
\\
\begin{subfigure}{0.13\textwidth}
\includegraphics[width=0.95\textwidth]{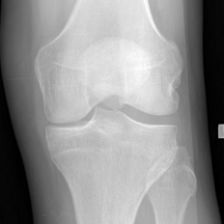}
\end{subfigure}
\begin{subfigure}{0.13\textwidth}
\includegraphics[width=0.95\textwidth]{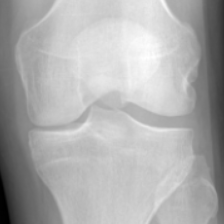}
\end{subfigure}
&
\begin{subfigure}{0.13\textwidth}
\includegraphics[width=0.95\textwidth]{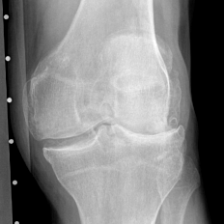}
\end{subfigure}
\begin{subfigure}{0.13\textwidth}
\includegraphics[width=0.95\textwidth]{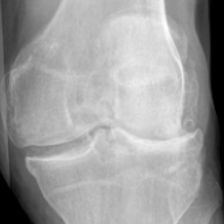}
\end{subfigure}
&
\begin{subfigure}{0.13\textwidth}
\includegraphics[width=0.95\textwidth]{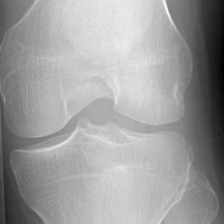}
\end{subfigure}
\begin{subfigure}{0.13\textwidth}
\includegraphics[width=0.95\textwidth]{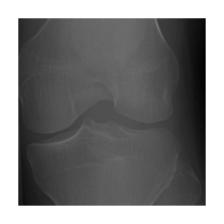}
\end{subfigure}
\\
\begin{subfigure}{0.13\textwidth}
\includegraphics[width=0.95\textwidth]{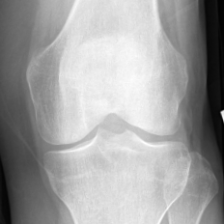}
\end{subfigure}
\begin{subfigure}{0.13\textwidth}
\includegraphics[width=0.95\textwidth]{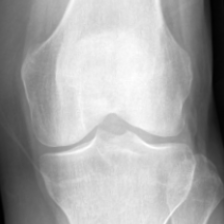}
\end{subfigure}
&
\begin{subfigure}{0.13\textwidth}
\includegraphics[width=0.95\textwidth]{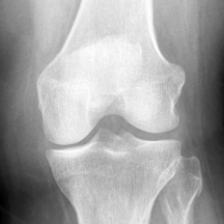}
\end{subfigure}
\begin{subfigure}{0.13\textwidth}
\includegraphics[width=0.95\textwidth]{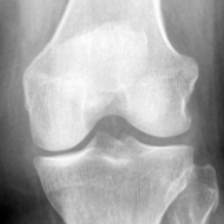}
\end{subfigure}
&
\begin{subfigure}{0.13\textwidth}
\includegraphics[width=0.95\textwidth]{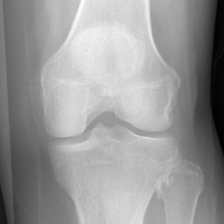}
\end{subfigure}
\begin{subfigure}{0.13\textwidth}
\includegraphics[width=0.95\textwidth]{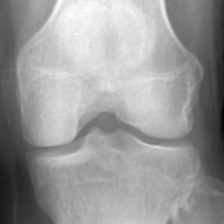}
\end{subfigure}
\\
\begin{subfigure}{0.13\textwidth}
\includegraphics[width=0.95\textwidth]{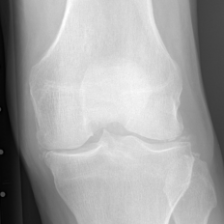}
\end{subfigure}
\begin{subfigure}{0.13\textwidth}
\includegraphics[width=0.95\textwidth]{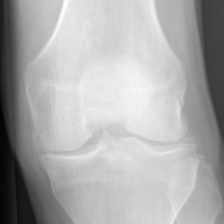}
\end{subfigure}
&
\begin{subfigure}{0.13\textwidth}
\includegraphics[width=0.95\textwidth]{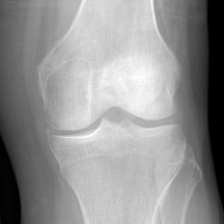}
\end{subfigure}
\begin{subfigure}{0.13\textwidth}
\includegraphics[width=0.95\textwidth]{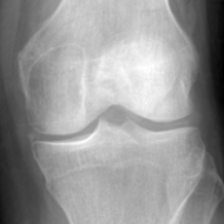}
\end{subfigure}
&
\begin{subfigure}{0.13\textwidth}
\includegraphics[width=0.95\textwidth]{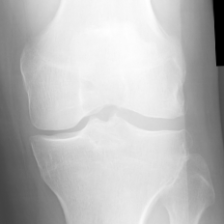}
\end{subfigure}
\begin{subfigure}{0.13\textwidth}
\includegraphics[width=0.95\textwidth]{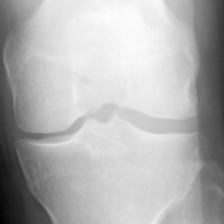}
\end{subfigure}
\\
\begin{subfigure}{0.13\textwidth}
\includegraphics[width=0.95\textwidth]{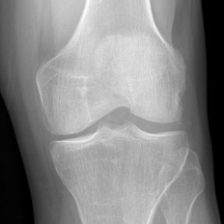}
\end{subfigure}
\begin{subfigure}{0.13\textwidth}
\includegraphics[width=0.95\textwidth]{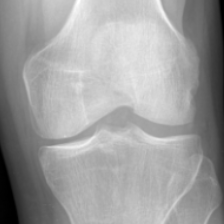}
\end{subfigure}
&
\begin{subfigure}{0.13\textwidth}
\includegraphics[width=0.95\textwidth]{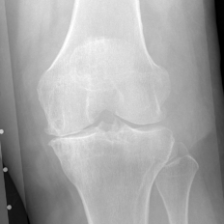}
\end{subfigure}
\begin{subfigure}{0.13\textwidth}
\includegraphics[width=0.95\textwidth]{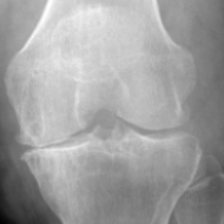}
\end{subfigure}
&
\begin{subfigure}{0.13\textwidth}
\includegraphics[width=0.95\textwidth]{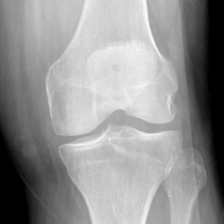}
\end{subfigure}
\begin{subfigure}{0.13\textwidth}
\includegraphics[width=0.95\textwidth]{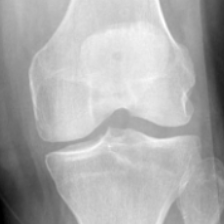}
\end{subfigure}
\\
\begin{subfigure}{0.13\textwidth}
\includegraphics[width=0.95\textwidth]{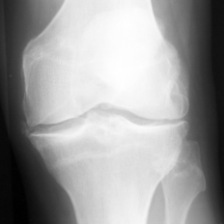}
\end{subfigure}
\begin{subfigure}{0.13\textwidth}
\includegraphics[width=0.95\textwidth]{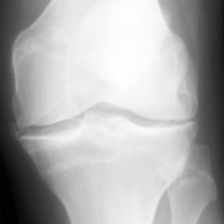}
\end{subfigure}
&
\begin{subfigure}{0.13\textwidth}
\includegraphics[width=0.95\textwidth]{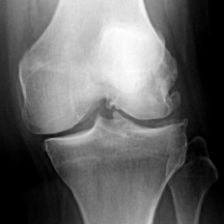}
\end{subfigure}
\begin{subfigure}{0.13\textwidth}
\includegraphics[width=0.95\textwidth]{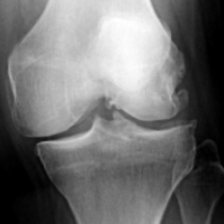}
\end{subfigure}
&
\begin{subfigure}{0.13\textwidth}
\includegraphics[width=0.95\textwidth]{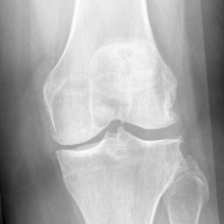}
\end{subfigure}
\begin{subfigure}{0.13\textwidth}
\includegraphics[width=0.95\textwidth]{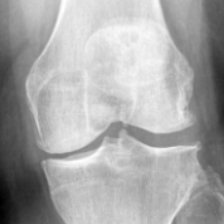}
\end{subfigure}
\end{tabular}
\caption{In each column separated by vertical bar, a left image is $u$ and the right image is $u(\pmb{\theta})$.}
\label{fig:num_results}
\end{figure*}

\begin{figure*}
\center
\begin{tabular}{c|c|c}
\begin{subfigure}{0.13\textwidth}
\includegraphics[width=0.95\textwidth]{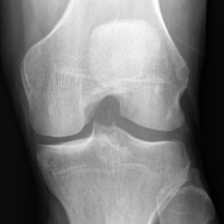}
\end{subfigure}
\begin{subfigure}{0.13\textwidth}
\includegraphics[width=0.95\textwidth]{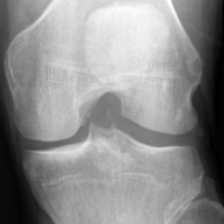}
\end{subfigure}
&
\begin{subfigure}{0.13\textwidth}
\includegraphics[width=0.95\textwidth]{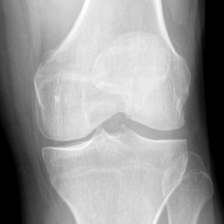}
\end{subfigure}
\begin{subfigure}{0.13\textwidth}
\includegraphics[width=0.95\textwidth]{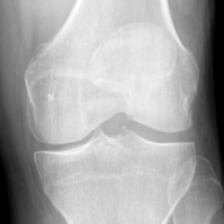}
\end{subfigure}
&
\begin{subfigure}{0.13\textwidth}
\includegraphics[width=0.95\textwidth]{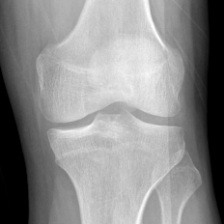}
\end{subfigure}
\begin{subfigure}{0.13\textwidth}
\includegraphics[width=0.95\textwidth]{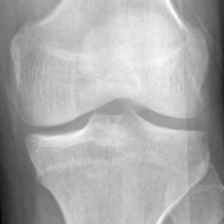}
\end{subfigure}
\\
\begin{subfigure}{0.13\textwidth}
\includegraphics[width=0.95\textwidth]{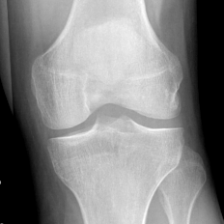}
\end{subfigure}
\begin{subfigure}{0.13\textwidth}
\includegraphics[width=0.95\textwidth]{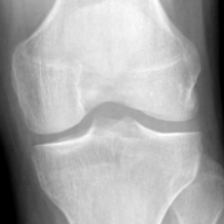}
\end{subfigure}
&
\begin{subfigure}{0.13\textwidth}
\includegraphics[width=0.95\textwidth]{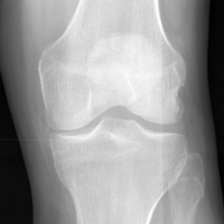}
\end{subfigure}
\begin{subfigure}{0.13\textwidth}
\includegraphics[width=0.95\textwidth]{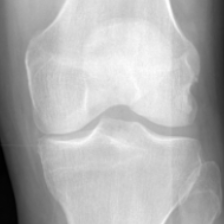}
\end{subfigure}
&
\begin{subfigure}{0.13\textwidth}
\includegraphics[width=0.95\textwidth]{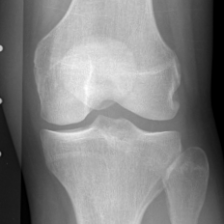}
\end{subfigure}
\begin{subfigure}{0.13\textwidth}
\includegraphics[width=0.95\textwidth]{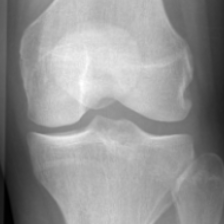}
\end{subfigure}
\\
\begin{subfigure}{0.13\textwidth}
\includegraphics[width=0.95\textwidth]{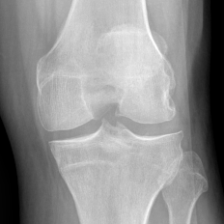}
\end{subfigure}
\begin{subfigure}{0.13\textwidth}
\includegraphics[width=0.95\textwidth]{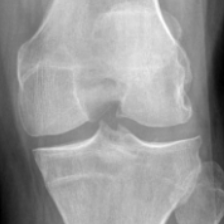}
\end{subfigure}
&
\begin{subfigure}{0.13\textwidth}
\includegraphics[width=0.95\textwidth]{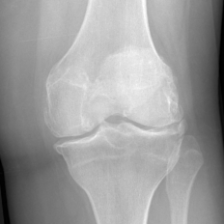}
\end{subfigure}
\begin{subfigure}{0.13\textwidth}
\includegraphics[width=0.95\textwidth]{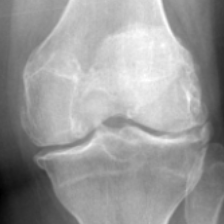}
\end{subfigure}
&
\begin{subfigure}{0.13\textwidth}
\includegraphics[width=0.95\textwidth]{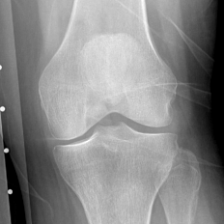}
\end{subfigure}
\begin{subfigure}{0.13\textwidth}
\includegraphics[width=0.95\textwidth]{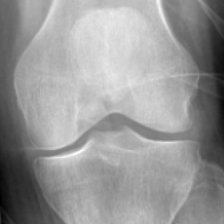}
\end{subfigure}
\\
\begin{subfigure}{0.13\textwidth}
\includegraphics[width=0.95\textwidth]{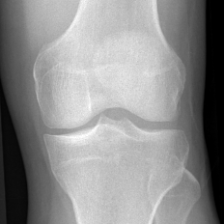}
\end{subfigure}
\begin{subfigure}{0.13\textwidth}
\includegraphics[width=0.95\textwidth]{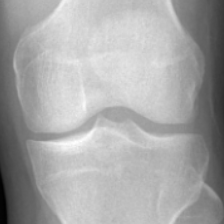}
\end{subfigure}
&
\begin{subfigure}{0.13\textwidth}
\includegraphics[width=0.95\textwidth]{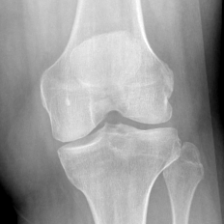}
\end{subfigure}
\begin{subfigure}{0.13\textwidth}
\includegraphics[width=0.95\textwidth]{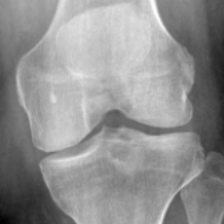}
\end{subfigure}
&
\begin{subfigure}{0.13\textwidth}
\includegraphics[width=0.95\textwidth]{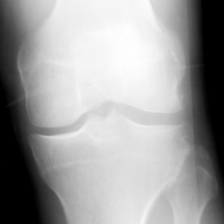}
\end{subfigure}
\begin{subfigure}{0.13\textwidth}
\includegraphics[width=0.95\textwidth]{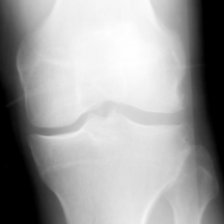}
\end{subfigure}
\\
\begin{subfigure}{0.13\textwidth}
\includegraphics[width=0.95\textwidth]{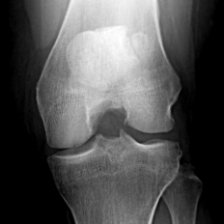}
\end{subfigure}
\begin{subfigure}{0.13\textwidth}
\includegraphics[width=0.95\textwidth]{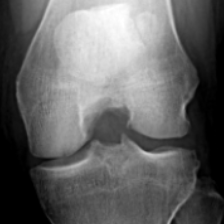}
\end{subfigure}
&
\begin{subfigure}{0.13\textwidth}
\includegraphics[width=0.95\textwidth]{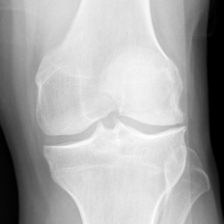}
\end{subfigure}
\begin{subfigure}{0.13\textwidth}
\includegraphics[width=0.95\textwidth]{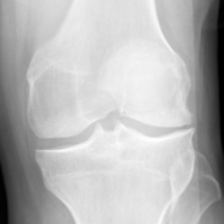}
\end{subfigure}
&
\begin{subfigure}{0.13\textwidth}
\includegraphics[width=0.95\textwidth]{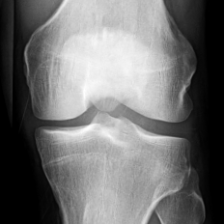}
\end{subfigure}
\begin{subfigure}{0.13\textwidth}
\includegraphics[width=0.95\textwidth]{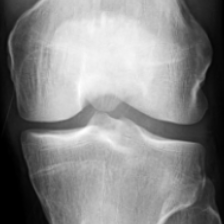}
\end{subfigure}
\\
\begin{subfigure}{0.13\textwidth}
\includegraphics[width=0.95\textwidth]{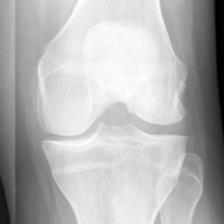}
\end{subfigure}
\begin{subfigure}{0.13\textwidth}
\includegraphics[width=0.95\textwidth]{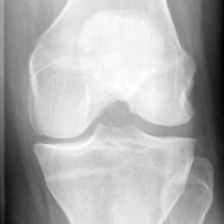}
\end{subfigure}
&
\begin{subfigure}{0.13\textwidth}
\includegraphics[width=0.95\textwidth]{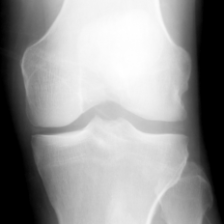}
\end{subfigure}
\begin{subfigure}{0.13\textwidth}
\includegraphics[width=0.95\textwidth]{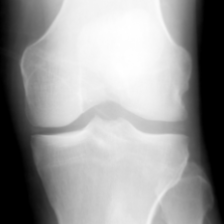}
\end{subfigure}
&
\begin{subfigure}{0.13\textwidth}
\includegraphics[width=0.95\textwidth]{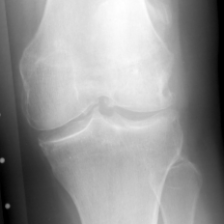}
\end{subfigure}
\begin{subfigure}{0.13\textwidth}
\includegraphics[width=0.95\textwidth]{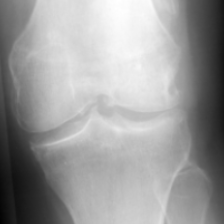}
\end{subfigure}
\\
\begin{subfigure}{0.13\textwidth}
\includegraphics[width=0.95\textwidth]{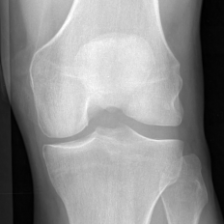}
\end{subfigure}
\begin{subfigure}{0.13\textwidth}
\includegraphics[width=0.95\textwidth]{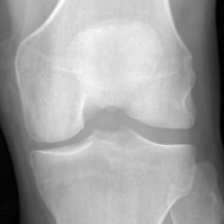}
\end{subfigure}
&
\begin{subfigure}{0.13\textwidth}
\includegraphics[width=0.95\textwidth]{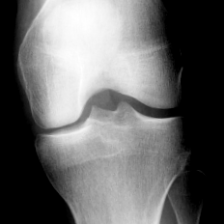}
\end{subfigure}
\begin{subfigure}{0.13\textwidth}
\includegraphics[width=0.95\textwidth]{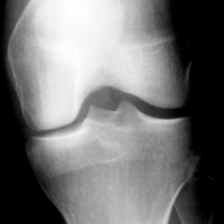}
\end{subfigure}
&
\begin{subfigure}{0.13\textwidth}
\includegraphics[width=0.95\textwidth]{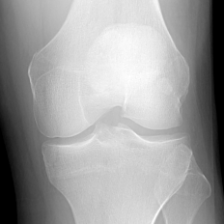}
\end{subfigure}
\begin{subfigure}{0.13\textwidth}
\includegraphics[width=0.95\textwidth]{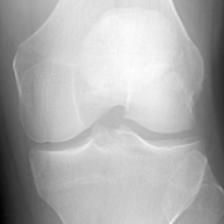}
\end{subfigure}
\\
\begin{subfigure}{0.13\textwidth}
\includegraphics[width=0.95\textwidth]{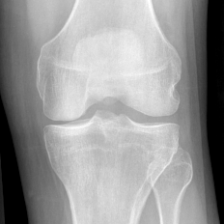}
\end{subfigure}
\begin{subfigure}{0.13\textwidth}
\includegraphics[width=0.95\textwidth]{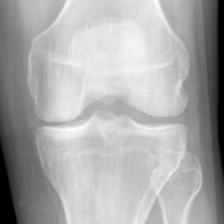}
\end{subfigure}
&
\begin{subfigure}{0.13\textwidth}
\includegraphics[width=0.95\textwidth]{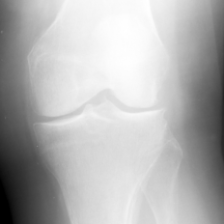}
\end{subfigure}
\begin{subfigure}{0.13\textwidth}
\includegraphics[width=0.95\textwidth]{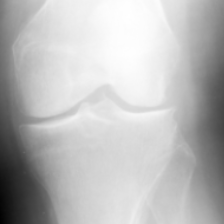}
\end{subfigure}
&
\begin{subfigure}{0.13\textwidth}
\includegraphics[width=0.95\textwidth]{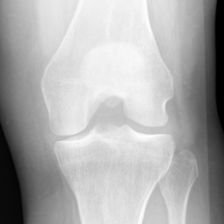}
\end{subfigure}
\begin{subfigure}{0.13\textwidth}
\includegraphics[width=0.95\textwidth]{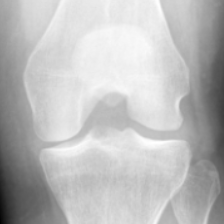}
\end{subfigure}
\\
\begin{subfigure}{0.13\textwidth}
\includegraphics[width=0.95\textwidth]{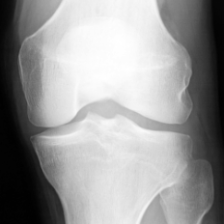}
\end{subfigure}
\begin{subfigure}{0.13\textwidth}
\includegraphics[width=0.95\textwidth]{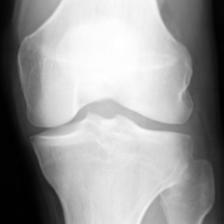}
\end{subfigure}
&
\begin{subfigure}{0.13\textwidth}
\includegraphics[width=0.95\textwidth]{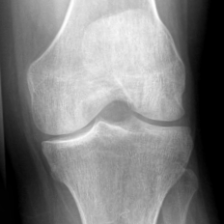}
\end{subfigure}
\begin{subfigure}{0.13\textwidth}
\includegraphics[width=0.95\textwidth]{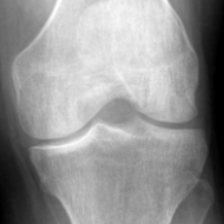}
\end{subfigure}
&
\begin{subfigure}{0.13\textwidth}
\includegraphics[width=0.95\textwidth]{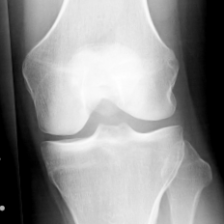}
\end{subfigure}
\begin{subfigure}{0.13\textwidth}
\includegraphics[width=0.95\textwidth]{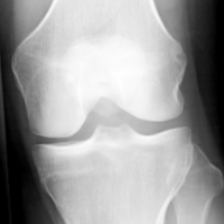}
\end{subfigure}
\\
\begin{subfigure}{0.13\textwidth}
\includegraphics[width=0.95\textwidth]{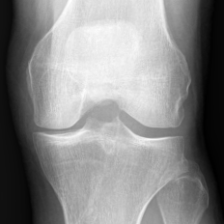}
\end{subfigure}
\begin{subfigure}{0.13\textwidth}
\includegraphics[width=0.95\textwidth]{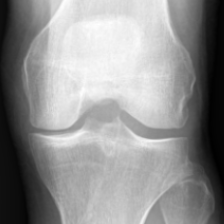}
\end{subfigure}
&
\begin{subfigure}{0.13\textwidth}
\includegraphics[width=0.95\textwidth]{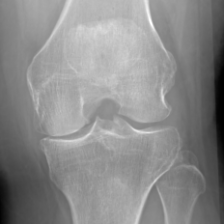}
\end{subfigure}
\begin{subfigure}{0.13\textwidth}
\includegraphics[width=0.95\textwidth]{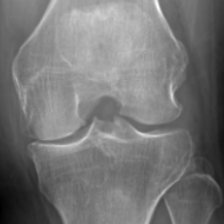}
\end{subfigure}
&
\begin{subfigure}{0.13\textwidth}
\includegraphics[width=0.95\textwidth]{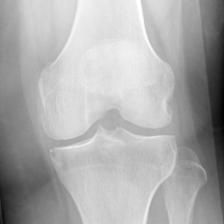}
\end{subfigure}
\begin{subfigure}{0.13\textwidth}
\includegraphics[width=0.95\textwidth]{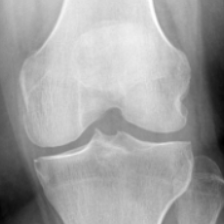}
\end{subfigure}
\\
\begin{subfigure}{0.13\textwidth}
\includegraphics[width=0.95\textwidth]{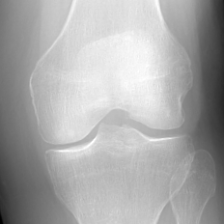}
\end{subfigure}
\begin{subfigure}{0.13\textwidth}
\includegraphics[width=0.95\textwidth]{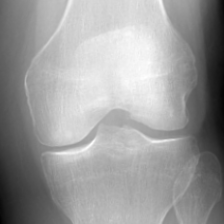}
\end{subfigure}
&
\begin{subfigure}{0.13\textwidth}
\includegraphics[width=0.95\textwidth]{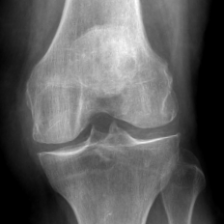}
\end{subfigure}
\begin{subfigure}{0.13\textwidth}
\includegraphics[width=0.95\textwidth]{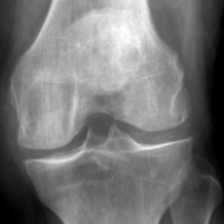}
\end{subfigure}
&
\begin{subfigure}{0.13\textwidth}
\includegraphics[width=0.95\textwidth]{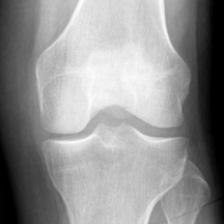}
\end{subfigure}
\begin{subfigure}{0.13\textwidth}
\includegraphics[width=0.95\textwidth]{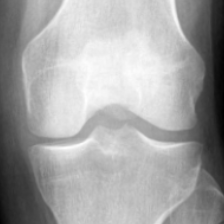}
\end{subfigure}
\end{tabular}
\caption{In each column separated by vertical bar, a left image is $u$ and the right image is $u(\pmb{\theta})$.}
\label{fig:num_results_2}
\end{figure*}

In Figures~\ref{fig:num_results} and \ref{fig:num_results_2} we see some numerical results. In the figures, there are three main columns separated by vertical bars and in each row of those columns there is a $(u,u(\pmb{\theta}))$ pair where $u$ is the neural network \cite{tiirola} output with some random added neighbourhood and $u(\pmb{\theta})$ is a candidate for the minimizer of \eqref{eq:u_test}. From the results we see that there is a closer resemblance between $u(\pmb{\theta})$ and the reference image $a$ than between $u$ and $a$. There are also less variations in sizes between the $u(\pmb{\theta})$ than between the $u$.
\section{Conclusion}
In this paper, a parts based loss was considered for finetune registering X-ray knee joint areas using a simple template image. The parts were first selected from a reference image. Then  a patch is extracted from a test image such that the detected parts in the patch have a similar spatial configuration than the parts in the reference image. In the experiments, it seemed visually that on average for input images roughly representing the knee joint areas the distance between the minimizers of the parts based loss function and the reference image is smaller than the distance between the input images and the reference image.
\bibliographystyle{abbrv}

\end{document}